# Automatically Restructuring Practice Guidelines using the GEM DTD

**Amanda Bouffier**          **Thierry Poibeau**
Laboratoire d'Informatique de Paris-Nord
Université Paris 13 and CNRS UMR 7030
99, av. J.-B. Clément – F-93430 Villetaneuse
`firstname.lastname@lipn.univ-paris13.fr`

## Abstract

This paper describes a system capable of semi-automatically filling an XML template from free texts in the clinical domain (practice guidelines). The XML template includes semantic information not explicitly encoded in the text (pairs of conditions and actions/recommendations). Therefore, there is a need to compute the exact scope of conditions over text sequences expressing the required actions. We present a system developed for this task. We show that it yields good performance when applied to the analysis of French practice guidelines.

## 1   Introduction

During the past years, clinical practices have considerably evolved towards standardization and effectiveness. A major improvement is the development of practice guidelines (Brownson *et al.*, 2003). However, even if widely distributed to hospitals, doctors and other medical staff, clinical practice guidelines are not routinely fully exploited[1]. There is now a general tendency to transfer these guidelines to electronic devices (via an appropriate XML format). This transfer is justified by the assumption that electronic documents are easier to browse than paper documents.

However, migrating a collection of texts to XML requires a lot of re-engineering. More precisely, it means analyzing the full set of textual documents so that they can fit with strict templates, as required either by XML schemas or DTD (document type definition). Unfortunately, most of the time, the semantic blocks of information required by the XML model are not explicitly marked in the original text. These blocks of information correspond to discourse structures.

This problem has thus renewed the interest for the recognition and management of discourse structures, especially for technical domains. In this study, we show how technical documents belonging to a certain domain (namely, clinical practice guidelines) can be semi-automatically structured using NLP techniques. Practice guidelines describe best practices with the aim of guiding decisions and criteria in specific areas of healthcare, as defined by an authoritative examination of current evidence (evidence-based medicine, see `Wikipedia` or Brownson *et al.*, 2003).

The Guideline Elements Model (GEM) is an XML-based guideline document model that can store and organize the heterogeneous information contained in practice guidelines (Schiffman, 2000). It is intended to facilitate translation of natural language guideline documents into a format that can be processed by computers. The main element of GEM, `knowledge component`, contains the most useful information, especially sequences of conditions and recommendations. Our aim is thus to format these documents which have been written manually without any precise model, according to the GEM DTD (see annex A).

The organization of the paper is as follows: first, we present the task and some previous approaches (section 2). We then describe the different processing steps (section 3) and the implementation (section 4). We finish with the presentation of some results (section 5), before the conclusion (section 6).

---

[1] See (Kolata, 2004). This newspaper article is a good example of the huge social impact of this research area.

## 2 Document Restructuring: the Case of Practice Guidelines

As we have previously seen, practice guidelines are not routinely fully exploited. One reason is that they are not easily accessible to doctors during consultation. Moreover, it can be difficult for the doctor to find relevant pieces of information from these guides, even if they are not very long. To overcome these problems, national health agencies try to promote the electronic distribution of these guidelines (so that a doctor could check recommendations directly from his computer).

### 2.1 Previous Work

Several attempts have already been made to improve the use of practice guidelines: for example knowledge-based diagnostic aids can be derived from them (*e.g.* Séroussi *et al.*, 2001).

GEM is an intermediate document model, between pure text (paper practice guidelines) and knowledge-based models like GLIF (Peleg *et al.*, 2000) or EON (Tu and Musen, 2001). GEM is thus an elegant solution, independent from any theory or formalisms, but compliant with other frameworks.

GEM Cutter ( http://gem.med.yale.edu/) is a tool aimed at aiding experts to fill the GEM DTD from texts. However, this software is only an interface allowing the end-user to perform the task through a time-consuming cut-and-paste process. The overall process described in Shiffman *et al.* (2004) is also largely manual, even if it is an attempt to automate and regularize the translation process.

The main problem in the automation of the translation process is to identify that a list of recommendations expressed over several sentences is under the scope of a specific condition (conditions may refer to a specific pathology, a specific kind of patients, temporal restrictions, etc.). However, previous approaches have been based on the analysis of isolated sentences. They do not compute the exact scope of conditional sequences (Georg and Jaulent, 2005): this part of the work still has to be done by hand.

Our automatic approach relies on work done in the field of discourse processing. As we have seen in the introduction, the most important sequences of text to be tagged correspond to discourse structures (conditions, actions …). Although most researchers agree that a better understanding of text structure and text coherence could help extract knowledge, descriptive frameworks like the one developed by Halliday and Hasan[2] are poorly formalized and difficult to apply in practice.

Some recent works have proposed more operational descriptions of discourse structures (Péry-Woodley, 1998). Several authors (Halliday and Matthiessen, 2004; Charolles, 2005) have investigated the use of non-lexical cues for discourse processing (*e.g* temporal adverbials like "*in 1999*"). These adverbials introduce situation frames in a narrative discourse, that is to say a 'period' in the text which is dependent from the adverbial.

We show in this study that condition sequences play the same role in practice guidelines: their scope may run over several dependent clauses (more precisely, over a set of several recommendations). Our plan is to automatically recognize these using surface cues and processing rules.

### 2.2 Our Approach

Our aim is to semi-automatically fill a GEM template from existing guidelines: the algorithm is fully automatic but the result needs to be validated by experts to yield adequate accuracy. Our system tries to compute the exact scope of conditional sequences. In this paper we apply it to the analysis of several French practice guidelines.

The main aim of the approach is to go from a textual document to a GEM based document, as shown on Figure 1 (see also annex A). We focus on conditions (including temporal restrictions) and recommendations since these elements are of paramount importance for the task. They are especially difficult to deal with since they require to accurately compute the scope of conditions.

The example on figure 1 is complex since it contains several levels of overlapping conditions. We observe a first opposition (*Chez le sujet non immunodéprimé / chez le sujet immunodéprimé... Concerning the non-immuno-depressed patient / Concerning the immuno-depressed patient...*) but a second condition interferes in the scope of this first level (*En cas d'aspect normal de la muqueuse iléale… If the ileal mucus seems normal...*). The task involves recognizing these various levels of conditions in the text and explicitly representing them through the GEM DTD.

---

[2] See "the text-forming component in the linguistic system" in Halliday and Hasan (1976:23).

⬇

```
<recommandation>
<decision.variable>Chez le sujet non immunodéprimé
</decsion.variable>
<decision.variable>en cas d'aspect macroscopique nor-
mal de la muqueuse colique </decison.variable>
<action> des biopsies coliques nombreuses et étagées
sont recommandées (...) </action>
<action>Les biopsies isolées sont insuffisantes(..)
</action>
<action>L'exploration de l'iléon terminal est égale-
ment recommandée</action>
</recommandation>

<recommandation>
<decision.variable>Chez le sujet non immunodéprimé
</decision.variable>
<decision.variable>en cas d'aspect macroscopique nor-
mal de la muqueuse colique </decision.variable>
<decision.variable>En cas d'aspect normal de la mu-
queuse iléale</decision.variable>
<action>la réalisation de biospsies n'est pas systéma-
tique</action>
</recommandation>

<recommandation
<decision.variable>Chez le sujet immunodépri-
mé</decision.variable>
<action> il est nécessaire de réaliser des biopsies
systématiques(...)</action>
</recommandation>
```

**Figure 1.** From the text to GEM

What is obtained in the end is a tree where the leaves are recommendations and the branching nodes correspond to the constraints on conditions.

## 2.3 Data

We analyzed 18 French practice guidelines published by French national health agency (ANAES, *Agence Nationale d'Accréditation et d'Evaluation en Santé* and AFSSAPS, *Agence Francaise de Sécurité Sanitaire des Produits de Santé)* between 2000 and 2005. These practice guidelines focus on different pathologies (e.g. diabetes, high blood pressure, asthma etc.) as well as with clinical examination processes (e.g. digestive endoscopy). amination processes (e.g. digestive endoscopy). The data are thus homogeneous, and is about 250 pages long (150,000+ words). Most of these practice guidelines are publicly available at: `http://www.anaes.fr` or `http://affsaps.sante.fr`. Similar documents have been published in English and other languages; the GEM DTD is language independent.

## 3 Processing Steps

Segmenting a guideline to fill an XML template is a complex process involving several steps. We describe here in detail the most important steps (mainly the way the scope of conditional sequences is computed), and will only give a brief overview of the pre-processing stages.

### 3.1 Overview

A manual study of several French practice guidelines revealed a number of trends in the data. We observed that there is a default structure in these guidelines that may help segmenting the text accurately. This default segmentation corresponds to a highly conventionalized writing style used in the document (a *norm*). For example, the location of conditions is especially important: if a condition occurs at the opening of a sequence (a paragraph, a section…), its scope is by default the entire following text sequence. If the condition is included in the sequence (inside a sentence), its default scope is restricted to the current sentence (Charolles, 2005 for similar observations on different text types).

This default segmentation can be revised if some linguistic cues suggest another more accurate segmentation (*violation of the norm*). We make use of Halliday's theory of text cohesion (Halliday and Hasan, 1976). According to this theory, some "cohesion cues" suggest extending the default segmentation while some others suggest limiting the scope of the conditional sequence (see section 3.4).

### 3.2 Pre-processing (Cue Identification)

The pre-processing stage concerns the analysis of relevant linguistic cues. These cues vary in nature: they can be based either on the material structure or the content of texts. We chose to mainly focus on task-independent knowledge so that the method is portable, as far as possible (we took inspiration from Halliday and Matthiessen's introduction to functional grammar, 2004). Some of these cues

(especially connectors and lexical cues) can be automatically captured by machine learning methods.

*Material structure cues*. These features include the recognition of titles, section, enumerations and paragraphs.

*Morpho-syntactic cues*. Recommendations are not expressed in the same way as conditions from a morpho-syntactic point of view. We take the following features into account:
- *Part of speech tags*. For example *recommandé* should be a verb and not a noun, even if the form is ambiguous in French;
- *Tense and mood of the verb*. Present and future tenses are relevant, as well as imperative and conditional moods. Imperative and future always have an injunctive value in the texts. Injunctive verbs (see *lexical* cues) lose their injunctive property when used in a past tense.

*Anaphoric cues*. A basic and local analysis of anaphoric elements is performed. We especially focused on expressions such as *dans ce cas, dans les N cas précédents* (*in this case, in the n preceding cases…*) which are very frequent in clinical documents. The recognition of such expressions is based on a limited set of possible nouns that occurred in context, together with specific constraints (use of demonstrative pronouns, etc).

*Conjunctive cues (discourse connectors)*. Conditions are mainly expressed through conjunctive cues. The following forms are especially interesting: forms prototypically expressing conditions (*si, en cas de, dans le cas où… if, in case of…*); Forms expressing the locations of some elements (*chez, en présence de... in presence of...*); Forms expressing a temporal frame (*lorsque, au moment où, avant de... when, before…*)

*Lexical cues*. Recommendations are mainly expressed through lexical cues. We have observed forms prototypically expressing recommendations (*recommander, prescrire, ... recommend, prescribe*), obligations (*devoir, ... shall*) or options (*pouvoir, ... can*). Most of these forms are highly ambiguous but can be automatically acquired from an annotated corpus. Some expressions from the medical domains can be automatically extracted using a terminology extractor (we use Yatea, see section 4, "Implementation").

### 3.3 Basic Segmentation

A *basic segment* corresponds to a text sequence expressing either a condition or a recommendation. It is most of the time a sentence, or a proposition inside a sentence.

Some of the features described in the previous section may be highly ambiguous. For this reason basic segmentation is rarely done according to a single feature, but most of the time according to a bundle of features acquired from a representative corpus. For example, if a text sequence contains an *injunctive* verb with an infinitive form at the beginning of a sentence, the whole sequence is typed as `action`. The relevant sets of co-occurring features are automatically derived from a set of annotated practice guidelines, using the chi-square test to calculate the dissimilarity of distributions.

After this step, the text is segmented into typed basic sequences expressing either a recommendation or a condition (the rest of the text is left untagged).

### 3.4 Computing Frames and Scopes

As for quantifiers, a conditional element may have a *scope* (a *frame*) that extends over several basic segments. It has been shown by several authors (Halliday and Matthiessen, 2004; Charolles, 2005) working on different types of texts that conditions detached from the sentence have most of the time a scope beyond the current sentence whereas conditions included in a sentence (but not in the beginning of a sentence) have a scope which is limited to the current sentence. Accordingly we propose a two-step strategy: 1) the default segmentation is done, and 2) a revision process is used to correct the main errors caused by the default segmentation (corresponding to the norm).

#### Default Segmentation

We propose a strategy which makes use of the notion of default. By default:

1. Scope of a heading goes up to the next heading;
2. Scope of an enumeration's header covers all the items of the enumeration ;
3. If a conditional sequence is detached (in the beginning of a paragraph or a sentence), its scope is the whole paragraph;
4. If the conditional sequence is included in a sentence, its scope is equal to the current sentence.

Cases 3 and 4 cover 50-80% of all the cases, depending on the practice guidelines used. However, this default segmentation is revised and modified when a linguistic cue is a continuation mark within the text or when the default segmentation seems to contradict some cohesion cue.

**Revising the Default Segmentation**

There are two cases which require revising the default segmentation: 1) when a cohesion mark indicates that the scope is larger than the default unit; 2) when a rupture mark indicates that the scope is smaller. We only have room for two examples, which, we hope, give a broad idea of this process.

1) Anaphoric relations are strong cues of text coherence: they usually indicate the continuation of a frame after the end of its default boundaries.

> **L'indication d'une insulinothérapie est recommandée lorsque l'HbA1c est > 8%, sur deux contrôles successifs sous l'association de sulfamides/metformine à posologie optimale.** Elle est laissée à l'appréciation par le clinicien du rapport bénéfices/inconvénients de l'insulinothérapie **lorsque l'HbA1c est comprise entre 6,6% et 8% sous la même association.** <u>Dans les deux cas</u>, la diététique aura au préalable été réévaluée et un facteur intercurrent de décompensation aura été recherchée (accord professionnel).
>
> Stratégie de prise en charge du patient diabétique de type 2 à l'exclusion de la prise en charge des complications (2000)

**Figure 2.** The last sentence introduced by *dans les deux cas* is under the scope of the conditions introduced by *lorsque*[3].

In Figure 2, the expression *dans les deux cas* (*in the two cases…*) is an anaphoric mark referring to the two previous utterances. The scope of the conditional segment introduced by *lorsque* (that would normally be limited to the sentence it appears in) is thus extended accordingly.

2) Other discourse cues are strong indicators that a frame must be closed before its default boundaries. These cues may indicate some contrastive, corrective or adversative information (*cependant, en revanche… however*). Justifications cues (*en effet, en fait … in effect*) also pertain to this class since a justification is not part of the `action` element of the GEM DTD.

Figure 3 is a typical example. The linguistic cue *en effet* (*in effect*) closes the frame introduced by

> **Chez les patients ayant initialement une concentration très élevée de LDL-cholestérol, et notamment chez les patients à haut risque dont la cible thérapeutique est basse (<1g/l),** le prescripteur doit garder à l'esprit que la prescription de statine à fortes doses ou en association nécessite une prise en compte au cas par cas du rapport bénéfice/risque et ne doit jamais être systématique. <u>En effet</u>, les fortes doses de statines et les bithérapies n'ont pas fait l'objet à ce jour d'une évaluation suffisante dans ces situations.
>
> (Prise en charge thérapeutique du patient dyslipidémique, 2005, p4)

**Figure 3.** The last sentence contains a justification cue (*en* effet) which limits the scope of the condition in the preceding sentence.

*Chez les patients ayant initialement...(<1g/l)* since this sequence should fill the `explanation` element of the GEM DTD and is not an `action` element.

## 4 Implementation

Accurate discourse processing requires a lot of information ranging from lexical cues to complex co-occurrence of different features. We chose to implement these in a classic blackboard architecture (Englemore and Morgan, 1988). The advantages of this architecture for our problem are easy to grasp: each linguistic phenomenon can be treated as an independent agent; inference rules can also be coded as specific agents, and a facilitator controls the overall process.

Basic linguistic information is collected by a set of modules called "linguistic experts". Each module is specialized in a specific phenomenon (text structure recognition, part-of-speech tagging, term spotting, *etc.*). The text structure and text formatting elements are recognized using Perl scripts. Linguistic elements are encoded in local grammars, mainly implemented as finite-state transducers (Unitex[4]). Other linguistic features are obtained using publicly available software packages, *e.g.* a part-of-speech tagger (Tree Tagger[5]) and a term extractor (Yatea[6]), *etc.* Each linguist expert is encapsulated and produces annotations that are stored in the database of facts, expressed in Prolog (we thus avoid the problem of overlapping XML tags, which are frequent at this stage). These annotations are indexed according to the textual clause they appear in, but linear ordering of the text is not cru-cial for further processing steps since the system mainly looks for co-occurrences of different cues. The resulting set of annotations constitutes the "working memory" of the system.

---

[3] In figures 2 and 3, bold and grey background are used only for sake of clarity; actual documents are made of text without any formatting.

[4] http://www-igm.univ-mlv.fr/~unitex/

[5] http://www.ims.uni-stuttgart.de/projekte/corplex/TreeTagger/DecisionTreeTagger.html

[6] http://www-lipn.univ-paris13.fr/~hamon/YaTeA

Another set of experts then combine the initial disseminated knowledge to recognize basic segments (section 3.3) and to compute scopes and frames (section 3.4). These experts form the "inference engine" which analyzes information stored in the working memory and adds new knowledge to the database. Even when linear order is irrelevant for the inference process new information is indexed with textual clauses, to enable the system to produce the original text along with annotation.

A facilitator helps to determine which expert has the most information needed to solve the problem. It is the facilitator that controls, for example, the application of default rules and the revision of the default segmentation. It controls the chalk, mediating among experts competing to write on the blackboard. Finally, an XML output is produced for the document, corresponding to a candidate GEM version of the document (no XML tags overlap in the output since we produce an instance of the GEM DTD; all potential remaining conflicts must have been solved by the supervisor). To achieve optimal accuracy this output is validated and possibly modified by domain experts.

## 5 Evaluation

The study is based on a corpus of 18 practice guidelines in French (several hundreds of frames), with the aid of domain experts. We evaluated the approach on a subset of the corpus that has not been used for training.

### 5.1 Evaluation Criteria

In our evaluation, a sequence is considered correct if the semantics of the sequence is preserved. For example *Chez l'obèse non diabétique (accord professionnel)* (*In the case of an obese person without any diabetes (professional approval)*), recognition is correct even if *professional approval* is not *stricto sensu* part of the condition. On the other hand, *Chez l'obèse* (*In the case of an obese person*) is incorrect. The same criteria are applied for recommendations.

We evaluate the scope of condition sequences by measuring whether each recommendation is linked with the appropriate condition sequence or not.

### 5.2 Manual Annotation and Inter-annotator Agreement

The data is evaluated against practice guidelines manually annotated by two annotators: a domain expert (a doctor) and a linguist. In order to evaluate inter-annotator agreement, conditions and actions are first extracted from the text. The task of the human annotators is then to (manually) build a tree, where each action has to be linked with a condition. The output can be represented as a set of couples (*condition − actions*). In the end, we calculate accuracy by comparing the outputs of the two annotators (# of common couples).

Inter-annotator agreement is high (157 nodes out of 162, i.e. above .96 agreement). This degree of agreement is encouraging. It differs from previous experiments, usually done using more heterogeneous data, for example, narrative texts. Temporals (like "*in 1999*") are known to open a frame but most of the time this frame has no clear boundary. Practice guidelines should lead to actions by the doctor and the scope of conditions needs to be clear in the text.

In our experiment, inter-annotator agreement is high, especially considering that we required an agreement between an expert and non-expert. We thus make the simplified assumption that the scope of conditions is expressed through linguistic cues which do not require, most of the time, domain-specific or expert knowledge. Yet the very few cases where the annotations were in disagreement were clearly due to a lack of domain knowledge by the non-expert.

### 5.3 Evaluation of the Automatic Recognition of Basic Sequences

The evaluation of basic segmentation gives the following results for the condition and the recommendation sequences. In the table, P is precision; R is recall; P&R is the harmonic mean of precision and recall (P&R = (2*P*R) / (P+R), corresponding to a F-measure with a $\beta$ factor equal to 1).

**Conditions:**

|     | Without domain knowledge | With domain knowledge |
| --- | --- | --- |
| P   | 1   | 1   |
| R   | .83 | .86 |
| P&R | .91 | .92 |

**Recommendations:**

|  | Without domain knowledge | With domain knowledge |
|---|---|---|
| P | 1 | 1 |
| R | .94 | .95 |
| P&R | .97 | .97 |

Results are high for both conditions and recommendations.

The benefit of domain knowledge is not evident from overall results. However, this information is useful for the tagging of titles corresponding to pathologies. For example, the title *Hypertension artérielle* (*high arterial blood pressure*) is equivalent to a condition introduced by *in case of...* It is thus important to recognize and tag it accurately, since further recommendations are under the scope of this condition. This cannot be done without domain-specific knowledge.

The number of titles differs significantly from one practice guideline to another. When the number is high, the impact on the performance can be strong. Also, when several recommendations are dependent on the same condition, the system may fail to recognize the whole set of recommendations.

Finally, we observed that not all conditions and recommendations have the same importance from a medical point of view – however, it is difficult to quantify this in the evaluation.

### 5.4 Evaluation of the Automatic Recognition of the Scope of Conditions

The scope of conditions is recognized with accuracy above .7 (we calculated this score using the same method as for inter-annotator agreement, see section 5.2).

This result is encouraging, especially considering the large number of parameters involved in discourse processing. In most of successful cases the scope of a condition is recognized by the default rule (default segmentation, see section 3.4). However, some important cases are solved due to the detection of cohesion or boundary cue (especially titles).

The system fails to recognize extended scopes (beyond the default boundary) when the cohesion marks correspond to lexical items which are related (synonyms, hyponyms or hypernyms) or to complex anaphora structures (nominal anaphora; hyponyms and hypernyms can be considered as a spe-cial case of nominal anaphora). Resolving these rarer complex cases would require "deep" domain knowledge which is difficult to implement using state-of-art techniques.

## 6 Conclusion

We have presented in this paper a system capable of performing automatic segmentation of clinical practice guidelines. Our aim was to automatically fill an XML DTD from textual input. The system is able to process complex discourse structures and to compute the scope of conditional segments spanning several propositions or sentences. We show that inter-annotator agreement is high for this task and that the system performs well compared to previous systems. Moreover, our system is the first one capable of resolving the scope of conditions over several recommendations.

As we have seen, discourse processing is difficult but fundamental for intelligent information access. We plan to apply our model to other languages and other kinds of texts in the future. The task requires at least adapting the linguistic components of our system (mainly the pre-processing stage). More generally, the portability of discourse-based systems across languages is a challenging area for the future.

**Annex A. Screenshots of the system**

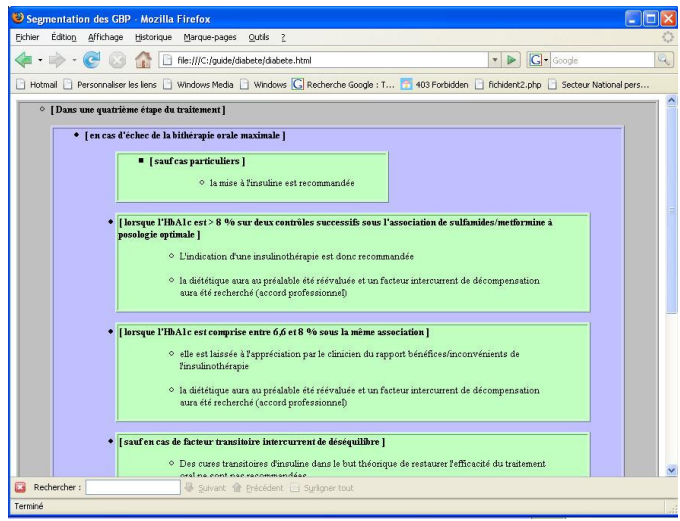

**Figure A1.** A practice guideline once analyzed by the system (*Traitement médicamenteux du diabète de type 2*, AFSSAPS-HAS, nov. 2006)

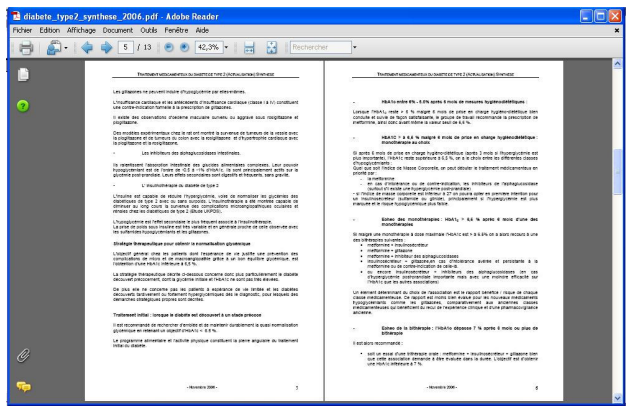

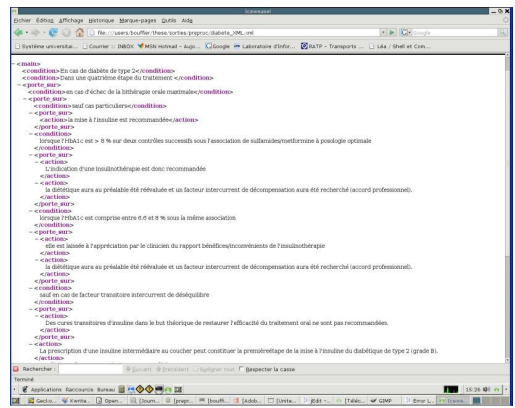

**Figures A2 and A3.** The original text, an XML GEM template instanciated from the text